# The Annotation Bottleneck in Persian Text NLP: Persian as an Annotation-Scarce Language

**MohammadHossein Mortazavi, Mostafa Salehi, Hadi Veisi**

Network Science and Technology Group, School of Intelligent Systems,
College of Interdisciplinary Science and Technologies, University of Tehran, Tehran, Iran
mh.mortazavi@ut.ac.ir, mostafa_salehi@ut.ac.ir, h.veisi@ut.ac.ir

**Abstract**

Persian (Farsi) is often described as a low-resource language in natural language processing, but that label collapses distinct shortages into a single category. This paper argues that Persian is more precisely described as annotation-scarce, provided that the term is understood as a property of its NLP resource ecology rather than an intrinsic property of the language. The review covers 34 representative Persian text resources available by July 2026 and adds three quantitative cross-checks. First, independent web measurements place Persian among roughly the twenty most visible content languages: W3Techs reports Persian on about 0.9% of websites with a known content language, while Common Crawl CC-MAIN-2026-30 identifies Persian as the primary language of 0.7039% of HTML pages. Second, a selective speech review shows a long resource trajectory from FARSDAT to recent corpora containing hundreds or thousands of hours of speech. Third, a matched Persian-English comparison normalizes task-specific annotation volumes by relative Common Crawl web presence. The resulting ratios vary sharply: Persian syntax and news NER are comparatively dense, whereas natural-language inference falls below the web-proportional baseline. The evidence therefore does not support a simple claim that Persian is globally deficient in labeled volume. Instead, annotation scarcity is expressed through uneven task and domain coverage, incompatible schemes, access and documentation friction, and limited supervision for specialist domains, preference data, and varieties beyond standard Iranian Persian.

## 1 Introduction

The term low-resource language has been useful in drawing attention to the unequal distribution of language technology. It becomes less informative when it is applied as a single language-level category. A language may have billions of unannotated tokens and still lack a reliable clinical named-entity test set. It may have a large part-of-speech corpus but no compatible discourse benchmark. Resources may also exist without being easy to reuse because their licenses, formats, annotation manuals, download locations, or version histories are incomplete. Taxonomies of linguistic resource availability and surveys of low-resource NLP already support a multidimensional interpretation [1], [2].

Persian is a particularly clear example. Its published resources include general news collections, web crawls, blogs, literary corpora, English-Persian parallel data, a six-million-pair colloquial-formal corpus, two sizeable Universal Dependencies treebanks, multiple named-entity and sentiment datasets, a 29,982-sentence proposition bank, discourse and coreference corpora, and a growing set of reasoning and language-understanding benchmarks. External web measurements point in the same direction: in late July 2026 W3Techs measured Persian on about 0.9% of websites with a known content language, and Common Crawl measured 0.7039% of HTML pages in CC-MAIN-2026-30 [3], [4]. These values place Persian within roughly the top twenty languages under two different web methodologies. This breadth demonstrates that Persian NLP cannot be assessed through raw pretraining volume alone. At the same time, merely counting datasets would overstate practical readiness. Many cover the same news, product-review, examination, or standard-written distributions, and their annotation schemes and access conditions are not uniform.

This paper asks whether annotation-scarce can serve as a more precise language-level description of Persian without becoming another absolute label. After defining the term and positioning the argument in prior work, the analysis combines three forms of evidence: a curated inventory of 34 Persian text resources, independent measurements of Persian web presence, and a selective speech-resource cross-check. It then distinguishes areas with substantial annotation from those where coverage remains narrow, difficult to access, or weakly standardized. Finally, it introduces a task-matched Persian-English comparison that relates annotation volume to relative web presence, and uses the resulting pattern to refine the annotation-scarce diagnosis. The paper is a structured review and position paper with a quantitative comparative component, not an exhaustive census or a controlled causal study.

# 2 Terminology and Scope

## 2.1 Annotation scarcity as a qualified language-level description

Joshi et al. classify languages by the amount and type of data available for NLP, while Hedderich et al. define low-resource scenarios in relation to insufficient labeled training data for a particular modeling setup [1], [2]. Both perspectives imply that resource status depends on the target task. This paper therefore uses annotation-scarce at two linked levels. At the task level, a task is annotation-scarce when available labels are too small, narrow, inaccessible, or weakly documented to support reliable training and evaluation. At the language-resource level, Persian is described as annotation-scarce when these shortages recur across enough important tasks, domains, and varieties to shape the practical state of the NLP ecosystem. The phrase is shorthand for the distribution and usability of resources; it is not an inherent property of Persian grammar, its speakers, or its digital presence, and it does not imply that every Persian NLP task lacks annotation.

Raw volume is only one component. A usable resource profile also depends on provenance, annotation quality, benchmark design, licensing, format, and coverage of the registers and varieties for which a system will be used. The intervention should follow the diagnosis. A missing raw corpus calls for collection and curation; a missing gold standard calls for annotation; an inaccessible corpus calls for licensing or distribution work; and an unreliable benchmark calls for redesign rather than simply more examples.

## 2.2 Language and modality scope

The main empirical inventory concerns text processing in standard Iranian Persian, written mainly in the Perso-Arabic script. Dari and Tajik are closely related standardized varieties, but they differ in sociolinguistic setting, vocabulary, orthographic practice, and, for Tajik, script. Pooling their speaker counts while reviewing mostly Iranian-Persian data would overstate the coverage of the resources discussed here. Section 5.5 adds a compact speech resource cross-check because speech data provide an independent test of the broader claim that Persian is not generally data-absent. That subsection is selective rather than a complete speech survey, and conclusions about automatic speech recognition, text-to-speech, speaker recognition, or prosodic annotation should remain task-specific.

# 3 Related Work

## 3.1 Low-resource classifications and dataset documentation

Research on low-resource NLP provides the conceptual starting point for this paper, but it does not by itself resolve how Persian should be described. Joshi et al. map languages according to the availability of data, tools, and research attention, showing that computational resource inequality is not reducible to speaker population [1]. Hedderich et al. approach the problem from the modeling side and define low-resource scenarios in relation to the labeled data available for a particular task and method [2]. These perspectives support a task-sensitive account, yet they operate at a broad multilingual or methodological level. They do not directly address the case of a language that has extensive raw corpora, established resources for several tasks, and recurring shortages of reusable annotation in other domains.

Work on dataset documentation adds a second foundation. Data statements and datasheets argue that a dataset cannot be evaluated only by its number of examples: provenance, collection procedure, population and language coverage, annotation process, intended use, licensing, and known limitations also determine whether it is scientifically reusable [5], [6]. This paper adopts that broader view of adequacy. Its use of annotation-scarce therefore concerns not only the absolute amount of labeled material, but also whether the material can be located, interpreted, combined, and applied to the task, domain, and language variety under study.

## 3.2 Persian corpus and annotation research

Persian NLP resources have largely developed through a sequence of corpus- and task-specific projects. Early work established reusable collections for information retrieval and morphosyntactic analysis, including Hamshahri, Peykare, the Uppsala Persian Corpus, and the dependency treebanks that were later represented in Universal Dependencies [7], [8]–[13]. Subsequent projects expanded the raw-text base through multi-source pretraining corpora, web crawls, blogs, literary collections, translation data, and colloquial–formal pairs [14]–[21]. This line of work demonstrates that Persian text exists at substantial scale and in more than one register or genre.

Speech-resource development provides a parallel history. Veisi's survey of Persian speech resources traces a progression from multilingual telephone corpora and FARSDAT to larger microphone, telephone, emotional, accented, child-speech, and text-to-speech collections [22]. More recent public resources substantially increase scale: DeepMine targets speaker verification and Persian ASR, Arman-AV adds audio-visual speech, Mozilla Common Voice supplies crowdsourced validated recordings, and ParsVoice targets large-scale multi-speaker TTS [23]–[26]. These resources reinforce the distinction between the amount of signal available and the depth of linguistic annotation attached to it.

A parallel line of research has produced supervision for particular tasks. PEYMA and ARMAN support named-entity recognition; SentiPers, MirasOpinion, ArmanEmo, and the Persian irony corpus support affective and pragmatic classification; ParsCORE addresses web register; and ParsiNLU, FarsTail, PerCQA, pn-summary, PersianMMLU, FarsEval-PKBETS, PARSE, and PersianPunc cover language understanding, inference, question answering, summarization, reasoning, and punctuation restoration [27]–[41]. Persian semantic and discourse research has also yielded PerPB, the Persian Discourse Treebank and coreference corpus, Mehr, a Persian RST corpus, and PerSemCor [42]–[46]. These contributions make claims that Persian has no annotated resources, or no semantic and discourse resources, untenable.

Most of these publications were designed to introduce a resource, document a construction procedure, or establish a benchmark for one task. Their local focus is appropriate, but it leaves a separate synthesis problem. Raw corpora, manually adjudicated labels, converted treebanks, crowdsourced annotations, automatically generated labels, and evaluation-only benchmarks represent different kinds of evidence. They also differ in domain, licensing, maintenance, tokenization, and compatibility. The present review does not replace the original dataset papers; it places their contributions in a shared resource profile so that breadth of publication is not mistaken for uniform practical adequacy.

### 3.3 Persian language models and evaluation

Model-centered research has made the distinction between text availability and annotation availability more visible. ParsBERT showed that a Persian-specific encoder could be pretrained on a heterogeneous monolingual corpus [14], while PersianMind and multilingual instruction-tuning efforts such as Aya demonstrate that Persian can also be represented in open generative-model development [47], [48]. At the evaluation level, ParsiNLU, PersianMMLU, FarsEval-PKBETS, and PARSE provide increasingly broad tests of understanding, knowledge, and reasoning [34], [38]–[40]. Comparative studies of general-purpose and open-source models further report uneven Persian performance across tasks and prompting conditions [49], [50].

These results establish that Persian model capability is neither absent nor uniform. They do not, however, identify annotation scarcity as the sole cause of every observed error. Performance can also depend on pretraining composition, tokenizer design, model scale, prompt language, benchmark construction, contamination, and domain shift. Model evaluation therefore supplies evidence about where systems fail, whereas a resource review is needed to determine whether the relevant training and gold evaluation data exist, are accessible, and match the target setting.

The literature reviewed above supplies the components of the present argument but not its full synthesis. Broad low-resource taxonomies do not provide a detailed Persian diagnosis; individual dataset papers do not assess the ecosystem as a whole; and model benchmarks do not by themselves distinguish raw-data limitations from annotation, access, interoperability, or coverage problems. This paper connects those strands. Its contribution is a structured, task-sensitive account of why Persian can possess many substantial datasets and still be described, with explicit qualifications, as an annotation-scarce language.

## 4 Resource Review Method

The text-resource snapshot was assembled from the ACL Anthology, Transactions of the Association for Computational Linguistics, Language Resources and Evaluation, LREC proceedings, arXiv, Universal Dependencies, and repositories linked by the corresponding publications. Searches combined Persian or Farsi with terms including corpus, treebank, parallel corpus, named entity recognition, sentiment, irony, web register, coreference, discourse, semantic role labeling, word sense, summarization, question answering, natural language inference, benchmark, normalization, and language model. Reference lists in major Persian resource papers were used to locate earlier datasets. The 34-resource text inventory is curated rather than exhaustive.

A resource was included when Persian was a target language and the publication reported enough information to identify its task, construction method, or scale. Model papers were included only when they introduced a corpus or benchmark, or supplied evidence directly relevant to resource adequacy. Distinct conversions of the same source corpus are listed separately only when the conversion changes interoperability or public access, as in the Universal Dependencies versions of Persian treebanks. Counts remain in the units reported by the source. Converting documents, tokens, sentences, and task instances into a single scale would create false precision.

The review favors publicly described resources with traceable documentation and covers general corpora, parallel data, linguistic annotation, classification datasets, and evaluation benchmarks. Publication does not guarantee that a dataset remains downloadable, has a clear license, or can be integrated without additional preprocessing. Commercial, unpublished, and institution-internal datasets may change the practical picture for particular organizations. The speech cross-check draws on Veisi's historical survey and on recent public corpus papers or datasheets [22]–[26], [51]; speech resources are not counted among the 34 text resources. AI-assisted tools were used to help locate candidate resources and cross-check reported statistics against their original publications.

Web presence was measured independently using W3Techs and Common Crawl. W3Techs estimates the share of websites using a content language among sites whose language it can identify; Common Crawl applies Compact

Language Detector 2 (CLD2) to HTML pages and reports the primary detected language [3], [4]. Their denominators and collection processes differ, so the percentages are not averaged or treated as token counts. Values are frozen to July 2026 to match the resource review.

For the Persian-English comparison, tasks were included only when a Persian resource and a well-documented English comparator could be expressed in the same source-reported unit. Tokens, sentences, documents, and task instances are not summed across tasks. Instead, Section 8.2 reports task-specific ratios normalized by the Persian-to-English Common Crawl share. This makes the comparison dimensionless while preserving task boundaries. The resulting index is illustrative and comparator-sensitive; it is not presented as a complete census of English annotation.

# 5 Raw Language Resources and Corpus Diversity

## 5.1 Persian in the global web ecosystem

Two independent web measurements place Persian far from the long tail of digitally marginal languages. W3Techs reported Persian on 0.9% of websites whose content language it knew on 26 July 2026; the same monthly survey placed English near 49.6%, Turkish at 1.6%, and Arabic at 0.6% [3]. W3Techs' historical series had Persian at 1.1% through late 2025 and early 2026 before falling below one percent, so an exact rank should always be paired with a date rather than treated as a permanent property of the language.

Common Crawl provides a different page-level measurement. In CC-MAIN-2026-30, CLD2 identified Persian as the primary language of 0.7039% of HTML pages, compared with 40.5782% for English, 1.3455% for Turkish, and 0.6548% for Arabic [4]. Persian is within the top twenty identified primary languages in that crawl. W3Techs counts websites, whereas Common Crawl counts crawled HTML pages; neither statistic is a direct token count. Their convergence nevertheless supplies independent evidence that Persian has a substantial digital footprint.

*Table 1. Web presence of selected languages in two independent July 2026 measurements. The measures have different denominators and should not be combined.*

| Language | W3Techs websites (%) | Common Crawl HTML pages (%) | Interpretive note |
|---|---|---|---|
| English | 49.6 | 40.5782 | Dominant baseline language |
| Turkish | 1.6 | 1.3455 | Regional comparison with larger web share |
| Persian | 0.9 | 0.7039 | Within roughly the top 20 under both views |
| Arabic | 0.6 | 0.6548 | Regional script-sharing comparison |

The quantitative consequence for this paper is straightforward. A language that occupies roughly seven to nine tenths of one percent of measured web content cannot be characterized simply as lacking digital text. The relevant question is whether labeled resources, evaluation data, and reusable annotation infrastructure have developed proportionally across the tasks in which Persian NLP is expected to operate.

## 5.2 General, web, and domain corpora

Persian text availability predates current large-language-model work. The Hamshahri collection contains 166,774 categorized newspaper documents and was designed as a standard information-retrieval test collection [7]. The ParsBERT study later assembled approximately 3.98 million documents from Wikipedia, news, magazines, travel and lifestyle sites, subtitles, and books, producing more than 38 million sentence-like segments [14]. MirasText broadened web-scale collection with roughly 2.84 million documents and 1.43 billion tokens drawn from more than 250 Persian websites [15].

The scale increased further with hmBlogs, which contains nearly 20 million blog posts and more than 6.8 billion tokens [16], and Matina, which reports 72.9 billion preprocessed and deduplicated tokens [17]. These corpora make a general claim of raw-text absence difficult to defend. They do not, however, solve questions of representativeness. News, blogs, and web pages can dominate a corpus while clinical notes, contracts, private dialogue, regional usage, and carefully documented colloquial text remain sparse.

Domain-specific collections add coverage that raw size alone cannot provide. The Corpus of Persian Literary Text spans the ninth to the twenty-first century, includes 112 poets and 57,980 poems or prose items, and contains partial rhetorical-figure annotation [18]. Its value lies less in competing with web corpora by token count than in supporting diachronic, stylistic, authorship, and literary-language research that ordinary news or crawl data cannot represent.

## 5.3 Parallel and variation-oriented corpora

Persian also has substantial parallel data. The Tehran English-Persian Parallel Corpus (TEP) provides 554,621 aligned subtitle lines, with approximately 3.71 million Persian words [19]. MIZAN contains 1,011,085 Persian-English sentence pairs drawn mainly from literary translations [20]. Their scale is useful for machine translation, but each has a strong domain signature: TEP reflects subtitle language and MIZAN reflects translated literature.

HarfoSokhan changes the picture for within-language variation. Released in 2026, it contains six million colloquial-formal Persian sentence pairs intended for style transfer and robustness to colloquial input [21]. This is a large resource by pair count, but its construction and task differ from independently written colloquial corpora. It should therefore be treated as a parallel variation resource, not as a complete description of naturally occurring spoken-style Persian.

*Table 2. Selected Persian corpora supporting general-text, domain, translation, and language-variation research. Counts are retained in the units reported by the sources.*

| Resource | Year | Reported scale | Primary material | Coverage / caveat |
|---|---|---|---|---|
| Hamshahri | 2009 | 166,774 documents; 564 MB | Categorized newspaper articles | Standard IR collection; 82 categories [7]. |
| ParsBERT pretraining corpus | 2020/2021 | ~3.98M documents; 38.27M segments | Wikipedia, news, magazines, subtitles, books | Multi-source model-training corpus; about 14 GB [14]. |
| MirasText | 2018 | ~2.84M documents; 1.43B tokens | Web text from >250 sites | Large automatically collected corpus; 15.3 GB reported [15]. |
| hmBlogs | 2021 | >6.8B tokens; nearly 20M posts | Persian blogs over ~15 years | Broad but blog-centered; raw and processed forms [16]. |
| Matina | 2025 | 72.9B tokens | Large web-derived corpus | Preprocessed and deduplicated; data and code reported public [17]. |
| Persian Literary Text | 2024 | 57,980 items; 9.14M poetry tokens | Poetry and prose, 9th–21st centuries | 112 poets; partial rhetorical-figure annotation [18]. |
| TEP | 2011 | 554,621 aligned lines | English-Persian movie subtitles | About 3.71M Persian words; colloquial/subtitle bias [19]. |
| MIZAN | 2018 | 1,011,085 sentence pairs | Persian-English literary translations | About 12.05M Persian words; literature-heavy [20]. |
| HarfoSokhan | 2026 | 6M sentence pairs | Colloquial-formal Persian | Large style-transfer resource; paired rather than independent text [21]. |

## 5.4 Interpreting raw-resource adequacy

Taken together, these corpora show that Persian has enough digital text to support substantial monolingual pretraining, information retrieval, translation, literary analysis, and colloquial-formal modeling. Raw-resource adequacy, however, cannot be inferred from volume alone. It also depends on which distributions are represented, whether the data can legally and technically be reused, and whether evaluation sets remain independent of model-training corpora. A billion web tokens and a million literary translation pairs serve different research needs and should not be collapsed into a single resource count.

## 5.5 Speech resources as complementary evidence

Although text remains the main scope of this article, Persian speech data provide a useful cross-check on the raw-data argument. The historical survey supplied by Veisi documents an established resource tradition beginning with multilingual OGI and CALLFRIEND data and the FARSDAT family [22]. FARSDAT itself contains roughly five hours of carefully recorded speech and about six thousand utterances with phonetic and phonemic segmentation; later projects expanded scale, including Large FARSDAT at roughly 140 hours and a telephone-conversation corpus at roughly 85 hours, of which only subsets received word-, phonetic-, or transcription-level annotation [22], [51]. That contrast between total audio and annotation depth is directly relevant to the distinction developed in this paper.

Recent resources have increased scale considerably. DeepMine contains more than 1,850 speakers and 540,000 recordings, with more than 480 hours transcribed across its Persian-English collection and dedicated Persian ASR protocols [23]. Arman-AV contributes almost 220 hours from 1,760 speakers for audio-visual speech tasks [24]. Mozilla Common Voice 26.0 reports 430.86 recorded hours of Persian, 373.24 validated hours, 394,397 clips, and 4,660 speakers [25]. ParsVoice processes 3,526 hours of audiobook speech and releases a 1,804-hour high-quality TTS subset with more than 470 speakers [26]. These corpora are not equivalent: some emphasize transcription, some validation, some audio-visual alignment, and some automated filtering for TTS.

*Table 3. Selected Persian speech resources, combining the historical survey with recent public corpora. This table is a complementary snapshot, not an exhaustive speech-resource census.*

| Resource | Year | Reported scale | Primary material | Annotation / caveat |
|---|---|---|---|---|
| FARSDAT | 1994 | ~5 h; ~6,000 utterances; ~300 speakers | Read microphone speech | Deep phonetic/phonemic segmentation [22], [51] |
| Large FARSDAT | early 2000s | ~140 h; 100 speakers | Read newspaper speech | Greater volume; microphone variation [22] |
| DeepMine | 2019 | >480 h transcribed overall; >1,850 speakers | Persian/English speech | Speaker verification + Persian ASR; bilingual total [23] |
| Arman-AV | 2024 | ~220 h; 1,760 speakers | Audio-visual Persian speech | ASR, lip reading, AV and speaker recognition [24] |
| Common Voice 26.0 | 2026 | 430.86 h recorded; 373.24 h validated; 4,660 speakers | Crowdsourced read speech | Open validated transcripts; not phonetic annotation [25] |
| ParsVoice | 2025 | 3,526 h processed; 1,804 h high-quality; >470 speakers | Audiobooks / TTS | Automated alignment and quality filtering [26] |

The speech evidence therefore points to the same distinction as the text evidence, but through a different modality. Persian speech is not generally absent at scale. What remains uneven is the type and depth of supervision: sentence transcripts, speaker metadata, phoneme boundaries, prosodic labels, spontaneous-speech coverage, dialect labels, and task-specific evaluation are distributed very differently across corpora. The annotation-scarce diagnosis should therefore not be read as a claim that Persian lacks audio; it concerns whether the annotation required by a particular speech or text task is available and reusable.

# 6 Annotated Datasets and Evaluation Resources

## 6.1 Morphosyntax and dependency parsing

Persian is comparatively well supplied for part-of-speech and syntactic annotation. Peykare contains approximately ten million words and remains one of the largest manually annotated written Persian corpora [8]. The Uppsala Persian Corpus (UPEC), a normalized and retokenized derivative of the Bijankhan corpus, contains 2,782,109 tokens with part-of-speech and morphological features [9]. The relationship between these legacy corpora matters: a large count does not imply compatibility with modern tokenization or dependency schemes.

For dependency parsing, Universal Dependencies release 2.18 provides two substantial treebanks. Persian-PerDT contains 29,107 sentences and 494,163 tokens, while Persian-Seraji contains 5,997 sentences and 151,627 tokens [10], [11]. Persian-PerDT derives from the treebank introduced by Rasooli et al. [12], whereas Persian-Seraji is associated with the conversion described by Seraji et al. [13]. Together they exceed 645,000 tokens, but their origins, genres, conversion histories, and tokenization decisions differ. Combining them without documenting those differences can create misleading gains or train-test overlap.

## 6.2 Entities, sentiment, pragmatics, and web register

Named-entity recognition has several established news-domain corpora. PEYMA contains 7,145 sentences and 302,530 tokens, including 41,148 entity-tagged tokens across seven classes [27]. ARMAN contains 7,682 sentences and 250,015 annotated tokens across six classes [28]. The NSURL-2019 shared-task dataset expands this base substantially: its training split contains 885,296 tokens and its test split 144,526 tokens, for 1,029,822 tokens in total across seven entity classes; the test set deliberately includes both in-domain and out-of-domain news [52]. The main limitation is therefore not absence of Persian NER supervision, and not even small news-domain volume. It is concentration in news, differences among label inventories and protocols, and limited coverage of medical, legal, scientific, conversational, privacy-oriented, and regional entity types.

Sentiment, emotion, and pragmatic annotation are broader than a single product-review dataset. SentiPers includes more than 26,000 review sentences with document-, sentence-, and aspect- or entity-level labels [29]. MirasOpinion contains 93,868 crowdsourced sentiment-labeled e-commerce documents [30]. ArmanEmo contributes more than 7,000 human-labeled social-media and e-commerce sentences across seven emotion classes [31], while the Persian irony corpus contains 4,339 manually labeled tweets [32]. These resources support meaningful work on opinion and affect, but their platform and topic distributions limit claims about political discourse, public services, literature, or regional usage.

ParsCORE adds a different layer of annotation: 2,000 Persian web documents labeled for online register or genre [33]. Register annotation is important because web-crawled corpora are not homogeneous. Distinguishing news, informational description, discussion, narrative, promotion, and other registers provides a more informative account of corpus composition than a single web-token total.

*Table 4. Selected Persian linguistic, sequence-labeling, semantic, discourse, and coreference resources.*

| Resource | Task | Reported scale | Construction | Coverage / caveat |
|---|---|---|---|---|
| Peykare | POS / written corpus | ~10M words | Manual; legacy tagset | Large broad corpus; mapping required [8]. |
| UPEC | POS + morphology | 2,782,109 tokens | Normalized derivative | Consistent segmentation; modified Bijankhan data [9]. |
| UD Persian-PerDT 2.18 | Dependency | 29,107 sent.; 494,163 tokens | Legacy treebank converted to UD | Multiple genres; conversion history matters [10], [12]. |
| UD Persian-Seraji 2.18 | Dependency | 5,997 sent.; 151,627 tokens | Manual + validated conversion | Multiple genres; CC BY-SA 4.0 [11], [13]. |
| PEYMA | NER | 7,145 sent.; 302,530 tokens | Manual | News; seven entity classes [27]. |
| ARMAN | NER | 7,682 sent.; 250,015 tokens | Manual | News; six entity classes [28]. |
| NSURL-2019 NER | NER | 1,029,822 tokens | Two-annotator shared-task corpus | News; seven classes; in/out-of-domain test [52] |
| ParsCORE | Web-register classification | 2,000 documents | Human annotation | Online genres/registers; emerging task [33]. |
| PerPB | Semantic roles | 29,982 sentences | Manual, double-checked stages | Verbs, propositional nouns and adjectives [42]. |
| Persian Discourse Treebank | Discourse + coreference | ~30K discourse sent.; 547 coref docs | Manual annotation | PDTB-style discourse and document coreference [43]. |
| Mehr | Coreference / pronouns | 400 documents | Manual corpus with resolution system | Useful but limited in scale and adoption [44]. |
| Persian RST Corpus | Discourse structure | 150 texts; ~400 words each | Manual RST annotation | Journalistic; 18 discourse relations [45]. |
| PerSemCor | Word-sense disambiguation | Automatically generated corpus | Automatic sense annotation | Silver resource, not independent gold data [46]. |

## 6.3 NLU, question answering, and summarization

Persian language-understanding datasets now cover several distinct task families. ParsiNLU introduced more than 14,500 instances across six tasks, including reading comprehension, textual entailment, question paraphrasing, sentiment analysis, multiple-choice question answering, and machine translation [34]. FarsTail contains 10,367 natural-language-inference examples generated from 3,539 multiple-choice questions [35].

PerCQA contains 989 community questions and 21,915 annotated answers for answer selection and ranking [36].

Summarization also has a sizeable resource. pn-summary contains 93,207 cleaned Persian news article-summary records from six news agencies [37]. The human-produced summaries make it useful for abstractive and extractive summarization, headline generation, and related tasks, although its news-source concentration should be retained in evaluation claims.

Recent evaluation suites have expanded rapidly. PersianMMLU contains 20,192 four-choice questions across 38 tasks [38]. FarsEval-PKBETS contains 4,000 questions from medical, legal, religious, linguistic, social, and general-knowledge categories [39]. PARSE provides 10,800 open-domain reasoning questions produced through a controlled generation and human-validation pipeline [40]. PersianPunc supplies 17 million constructed samples for punctuation restoration [41]. These resources are valuable, but question-answering benchmarks and derived training examples are not interchangeable with independently annotated token-level gold data.

*Table 5. Selected Persian classification, NLU, question-answering, summarization, and model-evaluation datasets.*

| Resource | Task | Reported scale | Construction | Coverage / caveat |
|---|---|---|---|---|
| SentiPers | Sentiment / ABSA | >26,000 sentences | Manual, multi-level | Digital-product reviews [29]. |
| MirasOpinion | Sentiment | 93,868 documents | Crowdsourced labels | E-commerce; class imbalance and domain concentration [30]. |
| ArmanEmo | Emotion | >7,000 sentences | Human labels | Social media and e-commerce; seven classes [31]. |
| Persian irony corpus | Irony detection | 4,339 tweets | Manual labels | Twitter-specific pragmatic dataset [32]. |
| ParsiNLU | Six NLU tasks | >14,500 instances | Expert/native annotation + curated material | Broad benchmark for standard Iranian Persian [34]. |
| FarsTail | NLI | 10,367 examples | Derived from 3,539 questions | Controlled NLI construction; possible artifacts [35]. |
| PerCQA | Community QA | 989 Q; 21,915 answers | Manual relevance annotation | Forum language; answer selection/ranking [36]. |
| pn-summary | Summarization | 93,207 records | News articles + human summaries | Six news agencies; source-domain concentration [37]. |
| PersianMMLU | Knowledge / reasoning | 20,192 questions; 38 tasks | Original Persian exam questions | Evaluation, not token-level training data [38]. |
| FarsEval-PKBETS | LLM evaluation | 4,000 questions | Curated benchmark | Medical, legal, religious, social and general domains [39]. |
| PARSE | Reasoning QA | 10,800 questions | Model generation + human validation | Boolean, multiple-choice and factoid formats [40]. |
| PersianPunc | Punctuation restoration | 17M samples | Aggregated and constructed | Large silver resource; not manual linguistic annotation [41]. |

## 6.4 Semantic roles, discourse, coreference, and lexical meaning

Persian semantic-role resources are not absent. The Persian Proposition Bank (PerPB) contains 29,982 manually annotated sentences, covers more than 9,200 unique verbs, and also annotates the argument structure of 1,300 propositional nouns and 300 propositional adjectives [42]. Its scale is substantial. The practical questions are current accessibility, format compatibility, licensing, and the extent to which later systems use a common train-test protocol.

Discourse and coreference annotation is also broader than one small corpus. The Persian Discourse Treebank reports 34,682 discourse relations over approximately 30,000 sentences, while its coreference component contains 547 documents, 212,646 tokens, 6,511 chains, and 21,303 mentions [43]. The Mehr corpus adds 400 documents for pronoun and coreference resolution [44]. A separate Persian RST corpus contains 150 journalistic texts, averaging roughly 400 words, annotated with 18 rhetorical relations [45]. These datasets demonstrate real annotation activity beyond the sentence, although their frameworks, availability, genres, and evaluation protocols are fragmented.

Lexical-semantic annotation is represented by PerSemCor, an automatically generated bag-of-words sense-annotated corpus intended for Persian word-sense disambiguation [46]. Because its labels are produced automatically, it should be distinguished from manually adjudicated gold annotation. Its inclusion is nevertheless important: weak and automatic annotation are part of the Persian resource landscape and can be useful when their provenance is explicit.

### 6.5 Annotation depth and practical availability

The tables distinguish manual or expert annotation, converted legacy annotation, curated benchmarks, crowdsourced labels, automatic sense labels, and model-generated data. These categories should not be added together as though they represented the same amount of human supervision. A converted dependency treebank preserves valuable manual work but can inherit conversion errors; a synthetic reasoning corpus can be useful for training but should not serve as an unquestioned independent test set; and an automatically sense-tagged corpus is not equivalent to an adjudicated semantic gold standard.

The number of published resources also exceeds the number of resources that a new researcher can necessarily retrieve and combine. Links disappear, licenses remain unspecified, preprocessing scripts depend on obsolete software, and related corpora may contain overlapping source texts. For Persian NLP, discoverability and maintenance are therefore part of the annotation bottleneck rather than secondary administrative concerns.

## 7 Linguistic and Technical Sources of Annotation Difficulty

### 7.1 Script normalization and tokenization

Persian text processing is affected by variation within the Perso-Arabic script. Visually similar but canonically distinct Unicode characters may be mixed across corpora, including Arabic and Persian forms of kaf and yeh. Graphemic-normalization studies show that these distinctions are not merely typographic: normalization can improve downstream language-modeling and machine-translation results [53]. The practical requirement is to document normalization rules rather than describe different code points as equally valid encodings of the same Persian character.

Spacing conventions create a second layer of variation. The zero-width non-joiner (ZWNJ) is used in standard forms such as “می‌روم” (mi-ravam, ‘I go’), but informal text may contain “میروم” or “می روم”. A tokenizer may treat these forms differently unless normalization and segmentation rules are explicit. The UD treebanks also differ in how clitics and multiword tokens are represented [10], [11]. Combining corpora without reconciling those decisions can introduce systematic label noise.

### 7.2 Morphology and ezafe

Persian morphology raises annotation questions, but broad claims such as “morphologically complex by most measures” are not informative. The relevant difficulties are specific: verbal prefixes and agreement suffixes, enclitic pronouns, plural and possessive morphology, light-verb constructions, and the ezafe linker. The short vowels that realize ezafe are usually not written after many word shapes, so a syntactic or speech-oriented annotation may need to recover information that is not overtly present in ordinary orthography [54].

*Table 6. Persian forms used to illustrate annotation decisions. The glosses are limited to the structure relevant to the discussion.*

| Written form | Transliteration | Relevant structure | Annotation issue |
|---|---|---|---|
| می‌روم | mi-rav-am | PROG-go.PRES-1SG | ZWNJ and morpheme boundaries |

| Written form | Transliteration | Relevant structure | Annotation issue |
|---|---|---|---|
| کتاب‌هایمان | ketāb-hā-ye-mān | book-PL-LINK-EZ-1PL.POSS | Four morphemes within one orthographic word |
| خانهٔ بزرگِ ما | xāne-ye bozorg-e mā | house-EZ big-EZ 1PL | Ezafe is partly or wholly unmarked in ordinary text |

### 7.3 Register, domain, and variety

Formal written Persian and colloquial Tehran Persian differ in pronouns, clitic placement, vocabulary, and orthographic practice. Social-media text adds code-switching, Latin-script Persian, emoji, nonstandard spacing, and creative spelling. HarfoSokhan and ParsCORE improve coverage of variation and register, but neither eliminates the need for independently sampled conversational data. Domain shift is equally important: entity inventories and syntactic patterns in medical notes or contracts differ from those in news and product reviews. Resource size alone therefore does not establish coverage.

## 8 A Task-Level and Cross-Language Diagnosis

### 8.1 Task-level resource profile

The evidence supports a differentiated diagnosis. Persian has substantial raw and parallel text, mature part-of-speech and dependency resources, multiple NER and affect datasets, and established resources for semantic and discourse analysis. The strongest remaining bottlenecks concern cross-resource compatibility, public access, documentation, specialist domains, independent evaluation, and representation beyond standard Iranian Persian. Taken together, this recurring pattern supports a qualified language-level description: Persian is annotation-scarce because shortages of reusable supervision affect multiple important parts of the NLP ecosystem, even though several individual tasks are comparatively well supplied.

*Table 7. Task-level resource profile for standard Iranian Persian text NLP. The categories are qualitative summaries, not universal thresholds.*

| Area | Current diagnosis | Evidence | Priority gap |
|---|---|---|---|
| General raw text | Substantial | Hamshahri, ParsBERT corpus, MirasText, hmBlogs, Matina | Provenance, licensing, deduplication, regional and specialist coverage |
| Parallel / variation data | Substantial but domain-skewed | TEP, MIZAN, HarfoSokhan | Subtitle/literary bias; independent colloquial corpora |
| POS / dependency | Substantial but fragmented | Peykare, UPEC, two UD treebanks | Tagset mapping, tokenization, cross-treebank evaluation |
| NER | Moderate to substantial in news; domain-limited | PEYMA, ARMAN, NSURL-2019 | Specialist domains, colloquial data, label compatibility |
| Sentiment / pragmatics | Moderate | SentiPers, MirasOpinion, ArmanEmo, irony corpus | Domain transfer, stance, political and regional discourse |
| General NLU / QA | Rapidly improving | ParsiNLU, FarsTail, PerCQA, PersianMMLU, FarsEval, PARSE | Contamination, construction artifacts, held-out evaluation |
| Summarization / generation | Moderate | pn-summary, PersianPunc | Non-news domains, gold evaluation, synthetic-data separation |
| Semantic roles | Substantial published resource | PerPB | Access, licensing, shared protocols, modern baselines |
| Discourse / coreference | Multiple resources, fragmented | Persian DT/coref, Mehr, Persian RST | Framework alignment, availability, cross-domain evaluation |
| Lexical semantics | Emerging / silver-heavy | PerSemCor | Manually adjudicated sense data and evaluation |
| Register and variety | Emerging | ParsCORE, HarfoSokhan | Regional Persian, natural dialogue, Latin-script Persian |
| Instruction / preference data | Poorly documented | Some multilingual community data may include Persian | Persian-specific provenance, annotator diversity, open licensing |

This framing also prevents unsupported causal conclusions. Dataset fragmentation, annotation cost, access constraints, and academic incentives are plausible contributors, but the present review does not measure their independent effects. Establishing those causes would require interviews, funding and publication analyses, or surveys of Persian NLP practitioners. They should not be stated as settled explanations on the basis of corpus counts alone.

## 8.2 Web-normalized annotation density relative to English

A single ratio formed by adding all labeled Persian samples and dividing by all unlabeled text would be numerically simple but scientifically misleading: one dependency token, one NLI pair, one coreference document, and one QA question do not represent the same unit of annotation. The comparison is therefore performed within tasks. Common Crawl supplies a shared raw-web baseline. In CC-MAIN-2026-30, the Persian-to-English page-share ratio is:

$$R_{web} = {}^{P_{fa}}\!\big/_{P_{en}} = {}^{0.7039}\!\big/_{40.5782} = 0.01735$$

For task $t$, let $A_{fa,t}$ and $A_{en,t}$ be the sizes of the selected Persian and English annotated resources measured in the same unit. The Web-Normalized Annotation Density ($WNAD$) is defined as:

$$WNAD_t = {}^{\left({}^{A_{fa,t}}\!\big/_{A_{en,t}}\right)}\!\Big/_{R_{web}}$$

$WNAD = 1$ means that the selected Persian annotation volume is proportional to Persian web presence relative to English. Values above one indicate a denser selected annotation resource than the web baseline; values below one indicate a thinner one. This is a resource-level diagnostic, not a universal quality score. English has many more datasets than any single comparator can represent, including multi-layer resources such as OntoNotes [55], so the table uses documented canonical resources and keeps the comparison task-specific.

***Table 8. Illustrative Persian-English task comparisons normalized by Common Crawl web presence. Ratios compare source-reported units within each task; WNAD is not an exhaustive ecosystem total.***

| Task | Persian annotated resource | English comparator | Fa/En | WNAD |
|---|---|---|---|---|
| Dependency parsing | 645,790 tokens: PerDT + Seraji [10], [11] | 1,102,940 tokens: 13 treebanks listed on English UD comparison page [56] | 0.586 | 33.8 |
| News NER | 1,029,822 tokens: NSURL-2019 [52] | 301,418 tokens: CoNLL-2003 English [52], [57] | 3.42 | 197.0 |
| Natural-language inference | 10,367 examples: FarsTail [35] | ~1.003M examples: SNLI + MultiNLI [58], [59] | 0.0103 | 0.60 |
| News summarization | 93,207 records: pn-summary [37] | 312,084 pairs: CNN/DailyMail standard splits [60] | 0.299 | 17.2 |

The comparison produces a mixed result rather than the expected monotonic deficit. Dependency parsing, the selected news NER resource, and news summarization all lie well above the web-proportional baseline, while FarsTail NLI lies below it. The spread is the substantive finding: Persian annotation is not uniformly scarce in aggregate volume. The very high news-NER value is also a warning about comparator sensitivity, because CoNLL-2003 is one canonical English benchmark rather than an inventory of all English NER data. For that reason no cross-task average is reported.

This quantitative result narrows the paper's central claim. Annotation-scarce should not mean that the total amount of Persian labeled data is small relative to its digital footprint. For several mature tasks it is not. The more defensible language-level diagnosis is coverage- and usability-based: annotation is abundant in particular islands of the ecosystem but repeatedly missing, domain-mismatched, incompatible, inaccessible, or weakly documented in other areas. The task-level profile in Table 7 and the normalized comparison in Table 8 should therefore be read together.

# 9 Large Language Models and the Annotation Bottleneck

Large multilingual and Persian-specialized language models change how annotation is used, but they do not remove the need for gold data. PersianMind extends a Llama-family model with Persian vocabulary and training on nearly two billion Persian tokens [47]. Its existence shows that Persian-specific open models are already part of the ecosystem. Multilingual instruction-tuning projects such as Aya likewise demonstrate how community-contributed instruction data can broaden language coverage [48].

Benchmark results nevertheless indicate uneven capability. PersianMMLU provides an original Persian evaluation set rather than a translated English benchmark [38]. Abaskohi et al. found that general-purpose models often remained behind Persian-fine-tuned systems on several tasks and that translating some Persian test items into English improved GPT-3.5 results [49]. FarsEval-PKBETS reported average accuracy below 50 percent for the three evaluated Persian-capable models [39]. A later open-source benchmark found that models struggled particularly with token-level NER relative to several sentence- or document-level tasks [50]. These results are compatible with an annotation bottleneck, but they do not isolate its contribution from pretraining volume, tokenizer design, model scale, prompting, or contamination.

LLMs also create new resource categories. Synthetic examples, model-proposed labels, and generated questions can accelerate data construction, as PARSE and PersianPunc illustrate [40], [41]. They should be accompanied by human validation, provenance records, and a clear separation between development and evaluation data. A large synthetic corpus may improve training while remaining unsuitable as an independent gold benchmark.

# 10 Discussion

The central question raised by this review is what it means to call Persian annotation-scarce when the language plainly has many datasets and a measurable global web presence. The web statistics make the first half of the diagnosis quantitative: Persian accounts for about 0.9% of known-language websites in a late-July W3Techs snapshot and 0.7039% of Common Crawl HTML pages in CC-MAIN-2026-30 [3], [4]. In the Common Crawl measurement, Persian web presence is about 1.735% of English. Researchers can therefore point not only to billions of Persian tokens but also to independent evidence that Persian is a non-marginal web language. At the same time, the task-matched comparison in Table 8 shows that labeled volume does not scale uniformly: selected Persian syntax, news NER, and summarization resources are large relative to that web baseline, whereas NLI is not. Resource abundance and resource scarcity coexist at different layers and in different tasks.

The speech cross-check leads to the same conclusion through another modality. Persian speech-resource development extends from the deeply segmented but small FARSDAT corpus to modern resources with hundreds of hours, and in the case of ParsVoice a high-quality TTS subset exceeding 1,800 hours [22]–[26], [51]. Yet hours alone do not say whether a researcher has phoneme boundaries, prosodic annotation, spontaneous conversational coverage, dialect labels, speaker metadata, or an independent test protocol. The raw-versus-annotation distinction is therefore not peculiar to text.

The most useful interpretation is therefore not that Persian lacks resources in general, nor that the existence of several dozen published datasets has resolved the problem. What persists is resource friction. The datasets reviewed here were created at different times, for different research questions, with different tokenization rules, label inventories, genres, licenses, and release practices. Taken as a whole, the landscape resembles a collection of valuable projects more than a coordinated research infrastructure. This is an interpretation of the documented pattern, not a claim about the intentions of the researchers who created those resources.

Three mismatches recur across the inventory. The first is between volume and annotation depth. Large web or blog corpora support representation learning, but they cannot replace task-specific labels for coreference, semantic roles, specialist entities, or human preferences. The second is between publication and practical usability. A corpus may be scientifically important while remaining difficult to retrieve, license, reproduce, or combine with newer resources. The third is between benchmark growth and training coverage. New question-answering and reasoning benchmarks reveal model failures, but they do not necessarily provide the gold token-level or domain-specific supervision needed to correct those failures. This helps explain why broad Persian language modeling can improve rapidly while specialized and cross-domain systems remain fragile.

The quantitative comparison also prevents annotation-scarce from becoming another blanket label. Calling Persian simply low-resource can encourage more collection and web crawling even where raw text is already plentiful. Calling it uniformly annotation-poor would now be equally inaccurate: the $WNAD$ values show that several mature Persian resources are large even after web normalization. Conversely, calling Persian well-resourced because many datasets can be named hides the fact that a news NER corpus does not solve clinical de-identification, a formal dependency treebank does not represent colloquial syntax, and an automatically constructed dataset is not an independent gold standard. Annotation-scarce is useful only as a coverage- and usability-based diagnosis of recurring gaps. The appropriate unit of analysis remains a task-domain-variety-resource profile.

The same point clarifies the role of large language models. Multilingual pretraining and Persian-specific models can exploit raw-text scale and reduce the amount of supervision needed for some applications. They do not make annotation irrelevant. Gold data remain necessary for measuring failure, separating linguistic weakness from prompting or model-scale effects, and refining systems in domains where generic pretraining is insufficient. As general model capability improves, the bottleneck shifts rather than disappears: from basic text availability toward reliable evaluation, specialist annotation, preference data, safety-relevant judgments, and coverage of colloquial and regional language.

This is why the paper was written. Its claim is not that Persian has little data, nor that its total labeled volume is uniformly small. The combined evidence instead supports a more specific diagnosis: Persian has a substantial raw digital presence and strong annotation islands, but reusable, well-documented, domain-appropriate supervision remains uneven enough to constrain work across multiple tasks and varieties. The value of annotation-scarce is therefore diagnostic rather than ordinal. It identifies where the next unit of effort should go – new human annotation, harmonization, licensing, maintenance, or better evaluation – without erasing the parts of Persian NLP that are already comparatively well resourced.

# 11 Recommendations

## 11.1 Maintain a versioned Persian resource registry

An immediately useful piece of infrastructure would be a public registry rather than another isolated list. Each entry should report task, variety, domain, size, annotation method, guideline availability, inter-annotator agreement, license, persistent identifier, current download status, version, and known overlap with other resources. Data statements and datasheet-style documentation provide established templates [5], [6]. The registry should distinguish original annotation from automatic conversion, weak labeling, and synthetic generation.

## 11.2 Prioritize coverage and interoperability gaps

The resource profile suggests that semantic-role and discourse resources do not need to be created from nothing; they need to be made easier to locate, license, convert, benchmark, and extend. New annotation should focus on multi-domain NER, clinical and legal text, natural conversation, regional varieties, stance, and robust cross-domain evaluation. For mature tasks, harmonization and carefully designed test sets may be more valuable than another dataset drawn from the same news distribution.

## 11.3 Make annotation quality auditable

Every manually labeled resource should release its guidelines or a detailed public summary, annotator training procedure, adjudication policy, and agreement statistics. Agreement should be interpreted with a metric appropriate to the task rather than reported as a single unqualified percentage [61]. Dataset splits, normalization rules, and tokenization versions should be fixed and versioned. Where copyright prevents redistribution, the reconstruction procedure and legal constraints should be stated plainly.

### 11.4 Use model-assisted annotation with safeguards

Active learning can reduce the number of examples that humans must label by selecting informative cases [62]. Modern models can also pre-annotate spans, propose relations, or flag disagreements. The efficiency gain is useful only if annotators can reject model suggestions and if the final evaluation set is independently checked. Error analysis should test whether model assistance amplifies weaknesses for rare entities, nonstandard spellings, minority varieties, or politically and culturally sensitive content.

### 11.5 Protect annotator welfare and data ethics

Persian datasets often contain social-media posts, reviews, forum questions, and culturally sensitive material. Future projects should document consent or lawful data use, privacy protection, compensation, annotator exposure to harmful content, and representation of regional and social groups. Annotation guidelines should allow disagreement where the task itself is subjective. A dataset can be large and still be scientifically weak if its annotators, provenance, or exclusions are invisible.

## 12 Limitations

This paper does not provide a complete census of Persian resources. It inventories 34 representative text resources and a smaller, selective set of speech corpora, while proprietary and institution-internal datasets remain outside the review. The text tables retain the measurement units used by source papers, which prevents a simple numerical ranking across tasks. Resource quality was assessed from published documentation rather than by re-annotating samples or reproducing every construction pipeline.

The inclusion of a publication also does not certify present-day usability. Some resources may have unavailable links, uncertain licenses, incomplete preprocessing code, or dependencies on older formats. The review separates construction types where documentation permits, but it does not independently verify every download, split, duplicate, or overlap. The 34-resource text inventory should therefore be read as a map of documented work, not a guarantee of frictionless access.

The main scope remains standard Iranian Persian text NLP. Section 5.5 is a selective speech cross-check rather than a speech census, and the paper still does not inventory Dari or Tajik resources. Web statistics introduce additional measurement limits: W3Techs counts websites, Common Crawl counts crawled HTML pages, language identification is automatic, and both distributions change over time [3], [4]. These measures are useful proxies for digital presence, not estimates of clean Persian training tokens.

The WNAD comparison is deliberately illustrative. It depends on the chosen English comparator, and canonical English datasets such as CoNLL-2003 do not represent the total amount of English annotation in existence. A high WNAD for one task therefore shows that a selected Persian resource is large relative to a matched benchmark and web baseline; it does not prove parity in domain coverage, quality, licensing, or the wider ecosystem. For the same reason, no single cross-task WNAD score is reported.

Finally, the paper does not establish why particular resource gaps arose. Funding, access to annotation platforms, institutional incentives, and geopolitical restrictions may matter, but causal claims about them require evidence beyond the resource counts reviewed here.

## 13 Conclusion

Persian can be described as an annotation-scarce language only in a qualified, coverage-based NLP sense. The phrase does not mean that Persian lacks digital text, audio, or substantial task-specific annotation. W3Techs and Common Crawl place Persian within roughly the top twenty web languages under two different methodologies [3], [4]; recent speech resources reach hundreds or thousands of hours [23]–[26]; and the published text landscape includes large general and domain corpora, substantial parallel data, mature morphosyntactic resources, multiple NER and affect datasets, a sizeable proposition bank, discourse and coreference annotation, and rapidly growing NLU and LLM benchmarks. The Persian-English comparison further shows that selected Persian annotation volumes can sit well above a web-proportional baseline for syntax, news NER, and summarization, while other tasks such as NLI remain thinner.

The annotation-scarce description should therefore function as a diagnostic property rather than a ranking of total labeled volume. Its value lies in identifying the recurring mismatch between substantial digital presence and the availability of reusable, domain-matched, interoperable, and well-documented supervision across the full task portfolio. The quantitative analysis makes that qualification necessary: Persian is not uniformly annotation-poor, but its resource ecology is uneven. Resource development should be planned at the level of tasks, domains, modalities, construction methods, accessibility, and language varieties. A versioned registry, interoperable standards, auditable annotation, and carefully governed model assistance would make existing work easier to reuse and direct new annotation toward gaps that are measurable rather than assumed.

### Acknowledgements

The authors thank colleagues, professors, and students at the Network Science and Technology Group, University of Tehran, for discussions that informed this paper. The work was prepared independently and without external financial support. Generative AI tools (specifically OpenAI's ChatGPT-4 and Anthropic's Claude Sonnet 5) were used during the preparation of this manuscript to improve the clarity and phrasing of author-drafted sentences. All AI-assisted text was reviewed, verified, and edited by the main author, who takes full responsibility for the final content.